\documentclass[11pt]{IEEEtran}
\IEEEoverridecommandlockouts

\usepackage{cite}
\usepackage{amsmath,amssymb,amsfonts}
\usepackage{algorithmic}
\usepackage{graphicx}
\usepackage{textcomp}
\usepackage{xcolor}
\usepackage{booktabs}
\usepackage{multirow}
\usepackage{url}
\usepackage{siunitx}

\usepackage{tikz}
\usepackage{pgfplots}
\pgfplotsset{compat = 1.6}
\usetikzlibrary{calc}

\def\BibTeX{{\rm B\kern-.05em{\sc i\kern-.025em b}\kern-.08em
    T\kern-.1667em\lower.7ex\hbox{E}\kern-.125emX}}
\begin{document}

\title{Region Masking to Accelerate Video Processing on Neuromorphic Hardware}
\author{Sreetama Sarkar$^{1}$ \ \ 
Sumit Bam Shrestha$^{2}$ \ \ 
Yue Che$^{1}$ \ \
Leobardo Campos-Macias$^{2}$ \\
Gourav Datta$^{3}$ \ \
Peter A. Beerel$^{1}$\\
$^{1}$Universiy of Southern California, Los Angeles, USA \ \ \ \ 
$^{2}$Intel Labs, Santa Clara, USA \\
$^{3}$Case Western Reserve University, USA \\
{\tt\small {\{sreetama,pabeerel,yueche\}@usc.edu} \ \  \tt\small {\{sumit.bam.shrestha,leobardo.e.campos.macias\}@intel.com} \ \ \tt\small {gourav.datta@case.edu}
}}
\maketitle

\begin{abstract}
The rapidly growing demand for on-chip edge intelligence on resource constrained devices has motivated approaches to reduce energy and latency of deep learning models. Spiking neural networks (SNNs) have gained particular interest due to their promise to reduce energy consumption using event-based processing. We assert that while sigma-delta encoding in SNNs can take advantage of the temporal redundancy across video frames, they still involve a significant amount of redundant computations due to processing insignificant events. In this paper, we propose a region masking strategy that identifies regions of interest at the input of the SNN, thereby eliminating computation and data movement for events arising from unimportant regions. Our approach demonstrates that masking regions at the input not only significantly reduces the overall spiking activity of the network, but also provides significant improvement in throughput and latency. We apply region masking during video object detection on Loihi 2, demonstrating that masking~$\sim$60\% of input regions can reduce energy-delay product by 
1.65{$\times$} over a baseline sigma-delta network, with a degradation in mAP@0.5 by~1.09\%.
\end{abstract}


\section{Introduction}
\label{sec:intro}
The widespread use of cameras in wearable and mobile devices has led to a rapidly growing demand for on-chip video processing. Artificial Neural Networks (ANNs) including Convolutional neural networks (CNNs) \cite{he2016deep} and Vision Transformers (ViTs) \cite{dosovitskiy2020image} have shown remarkable performance in computer vision tasks, including image recognition \cite{russakovsky2015ImageNet} and object detection \cite{lin2014microsoftcoco}. However, their high computational demand creates challenges for edge deployment. This has led to the growth of brain-inspired Spiking Neural Networks (SNNs) \cite{neuro_frontiers, dsnn_conversion1} which promises to significantly reduce energy consumption using event-based processing. The benefits of SNNs have been effectively realized in the real world with the development of neuromorphic processors, including Intel Loihi \cite{davies2018loihi}, SynSense \cite{synsense}, and SpiNNaker \cite{furber2020spinnaker}.

Several efforts \cite{kundu2021spike, rathi2018stdp} have been made to reduce the spiking activity and, thereby, the energy consumption in SNNs. 
Researchers have extensively explored pruning \cite{rathi2018stdp, shi2019soft} and quantization \cite{lu2020exploring, sorbaro2020optimizing} techniques to reduce the computational cost of SNNs. 
Sigma-delta encoding \cite{shrestha2024efficient} in SNNs exploits the temporal redundancy at both the input and intermediate layers by processing only the information that changed from the last timestep. However, SNNs still involve significant redundant computations due to processing events irrespective of their significance to a particular task. In this paper, we propose a region masking strategy that eliminates events arising from unimportant regions in an image, thereby reducing unnecessary computation and data movement.  

\noindent\textbf{Our Contributions:} 
In this paper, we propose a hardware-aware \textbf{\textit{input region masking strategy for accelerating video inference on SNNs deployed on neuromorphic hardware}}. Our approach leverages a combination of static and dynamic region masks, where sparsity levels are \textbf{\textit{carefully configured to minimize communication overhead while maintaining accuracy}}. The dynamic mask is generated using a self-attention-based region mask generator, while the static mask captures object locations from the training set. We evaluate our approach on three different video object detection datasets: Kitti \cite{Geiger2012CVPR}, BDD100K \cite{Yu_2020_CVPR} and, ImageNet-VID \cite{russakovsky2015ImageNet}, and demonstrate results on Intel Loihi 2 \cite{orchard2021loihi2}. Our region masking approach significantly reduces the overall spiking activity in the network, improving throughput by 1.22$\times$, and overall energy-delay product by 1.65$\times$ with accuracy degradation of~$1.09\%$ over baseline SNN detection.

\section{Related Work}
\label{sec:related_work}
\noindent\textbf{Spiking Neural Networks:}
Spiking Neural Networks (SNNs) compute and communicate using binary spikes distributed over several time steps, which are typically sparse and require only accumulation operations in convolutional and linear layers \cite{neuro_frontiers,dsnn_conversion1}. This results in significant computational efficiency. Although SNNs traditionally utilized binary spikes, modern neuromorphic chips now employ multi-bit graded spikes to enhance performance \cite{panda_res,spiking_lstm}. Historically, training SNNs posed challenges due to the zero gradients of the spike activation function \cite{panda_res,spiking_lstm}. However, advancements such as backpropagation through time (BPTT) \cite{lee_dsnn} and the integration of state-of-the-art Artificial Neural Network (ANN) training techniques through ANN-to-SNN conversion \cite{Rathi2020Enabling} have addressed these issues. Recently, SNNs have achieved competitive results in complex vision tasks like object detection and semantic segmentation using only a few time steps \cite{su2023deep,kim2021classificationdirectlytrainingspiking}.

\noindent\textbf{Loihi 2:} Loihi 2 is the second generation neuromorphic research chip from Intel. It is a fully event-driven digital asynchronous interconnect of neuromorphic cores that implement a variety of synaptic interconnect including convolution, fully programmable spiking neuron dynamics, and integer valued spikes (graded spikes). Loihi 2 is fully equipped to leverage the sparse messaging and event-driven computation pattern prevalent in SNNs and thus enable extremely low power and low latency solutions using sparse networks such as SNNs. A single Loihi 2 chip consists of 120 neuro cores as well as dedicated IO and embedded cores which can work in isolation as well as a cluster of thousands of chips which form the world's largest neuromorphic system~\cite{intelHalaPoint}. 

\noindent\textbf{Region Masking:} Masking or dropping input regions or patches have been widely adopted for accelerating inference, particularly in ViTs \cite{liang2022evit, fayyaz2022ats, bolya2022tokenmerge}. This is because, due to patch-based processing in ViTs, dropping input regions can yield benefits even on general-purpose hardware like GPUs. Region masking approaches have also been applied to CNNs, where the backend network either only processes regions of interest (RoI) \cite{feng2022eye}, or adopt a multi-resolution setting processing RoI at a higher resolution \cite{reidy2024hirise}, and the rest at a lower resolution. However, determining the RoI in these methods~\cite{feng2022eye, reidy2024hirise, sarkar2024maskvd} rely on the output of the model for previous frames in the video sequence. This may lead to an accumulation of errors and is not suitable particularly for small networks used for edge deployment. In this paper, we adopt a region masking strategy that is independent of the output of previous frames. In SNNs, energy reduction can be obtained by reducing spiking activity through spike masking. Masked Spiking Transformer \cite{Wang_2023_ICCV_masked_spikformer} introduces random spike masking to reduce power consumption, however, random masking severely degrades performance on complicated tasks. 
\section{Proposed Approach}
\label{sec:approach}


\subsection{Region Masking:}
We propose an input region masking strategy that creates input masks by combining a static mask, derived from object locations in the training set, with a dynamic mask, generated by a region mask generator network based on the current input frame. The static mask helps ensure some consistency in removal of the same regions across frames, while the dynamic mask adapts to the current frame, ensuring that newly appearing objects are captured.

\noindent\textbf{Static Mask:} We create the static mask using object locations from the training set, following the method in \cite{sarkar2024maskvd}. Initially, an accumulated heatmap \si{H} is generated from the ground-truth bounding boxes of training images. For each image in the training set, pixels containing objects are marked with 1's, and all others with 0's, producing a binary map. These binary maps are accumulated together to produce \si{H}. The pixel-wise values in \si{H} are then aggregated to compute region-wise scores for regions of size \si{p}$\times$\si{p} pixels. Based on the static keep rate ($k_{s}$), the top-$k$ regions are selected according to these scores to generate the static mask {Static Mask} ($Mask_{stat}$). $Mask_{stat}$ promotes consistency in the removal of the same regions across frames, reducing the need to process additional regions and enhancing energy efficiency.

\noindent\textbf{Dynamic Mask:}
The static mask alone is inadequate as it poses safety risks by potentially missing objects in the current frame. On the other hand, relying on the output from the previous frame increases the chance of error propagation, especially in smaller models. To address this, we generate a dynamic mask using a lightweight region \underline{M}ask \underline{G}enerator \underline{Net}work (MGNet) \cite{kaiser2024energy} that accurately predicts RoI based on the current frame. This network consists of a single transformer block followed by a self-attention and linear layer. MGNet decomposes input images into \si{p}$\times$\si{p} non-overlapping patches and maps each input patch into an embedding vector of length $L$. The embedding vectors are then passed into the transformer block and the self-attention layer, generating attention scores for each patch. MGNet utilizes the attention score, $\si{S}_{cls\_attn}$, computed as the dot product between the query vector derived from the $cls\_token$ ($\si{q}_{class}$) and the key matrix \si{K} from the other patches. 

\vspace{-4mm}
\begin{equation}
    {\si{S}_{cls\_attn}} = \frac{\si{q}_{class} \si{K}^T}{\sqrt{d}}.
    \label{eq:stoken}
\end{equation}

$\si{S}_{cls\_attn}$ inherently captures the importance of each patch. This attention score is fed into a linear layer of the same output dimension as the number of image patches to generate the importance scores $\si{S}_{region}$ for each region in the input image. The region scores are then passed through a $Sigmoid$ function and thresholded using a region threshold $t_{reg}$ to produce a 2D array of binary values, which we refer to as our {Dynamic Mask} ($Mask_{dyn}$). 

\begin{figure*}
    \centering
    \vspace{-8mm}
    \includegraphics[width=0.7\linewidth]{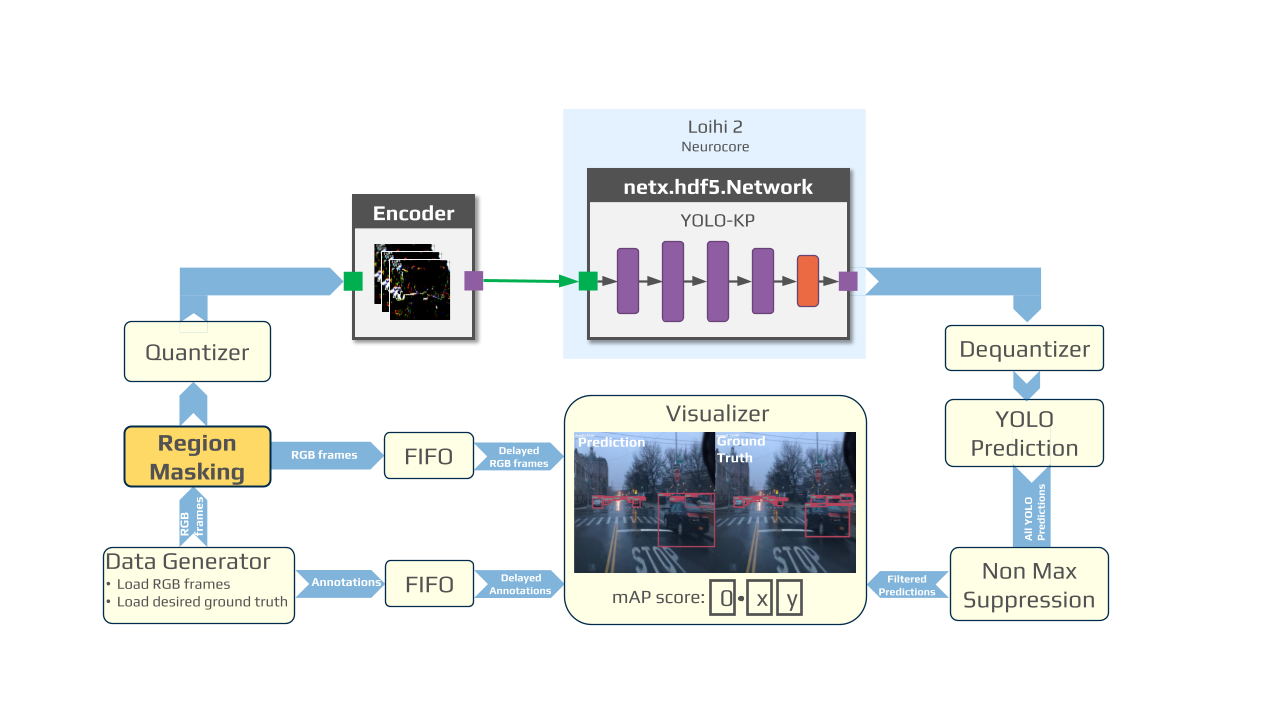}
    \vspace{-8mm}
    \caption{Inference pipeline on Loihi 2 using region masking. The Region Masking module applies masking to the RGB frames generated by the Data Generator and transmits the masked frame to the Quantizer module for further processing.}
    \label{fig:loihi}
\end{figure*}

MGNet is trained using binary cross-entropy loss between the predicted region scores $\si{S}_{region}$ and the ground truth labels obtained from the ground truth bounding boxes for that frame. In the ground truth label, a region is marked as \textit{one} if it contains an object fully or partially, and is marked as \textit{zero} otherwise. 

The input region mask is constructed using a union of the regions present in $Mask_{stat}$ and $Mask_{dyn}$. The mask contains binary values for each \si{p}$\times$\si{p} region, with \textit{ones} indicating regions to be processed and \textit{zeros} indicating regions to be skipped.


\subsection{Masked inference on Loihi 2}\label{subsec:loihi_inference}
\noindent\textbf{ANN model:} A Loihi 2 compatible Tiny-YOLOv3 inspired architecture (YOLO-KP) trained on the target dataset is used as the baseline model. The model is then fine-tuned with a static input mask described above to be compatible with masked inputs. The ANN is quantized to integer precision.

\noindent\textbf{Sigma-Delta conversion:} We use a Sigma-Delta pair between each layer to sparsify the spike messages between layers. In the sender layer,  delta encoding logic computes the change at each pixel and only spikes when the change is large enough to cross a threshold, reducing redundant information and increasing the sparsity of spike messages. For a temporal signal $x[t]$, the sigma spike message, $s[t]$, is described by
\begin{equation}
\begin{aligned}
    s[t] &= (x[t] - x_\text{ref}[t - 1])\,\mathcal{H}(|x[t] - x_\text{ref}[t - 1]| - \vartheta)\\
    x_\text{ref}[t] &= x_\text{ref}[t - 1] + s[t]
\end{aligned}
\end{equation}
where $x_\text{ref}$ is the reference state representing the previous spike that was sent and $\vartheta$ is the threshold.
The sparse spike messages in-turn give rise to sparse synaptic computation, thus taking advantage of the event based neuromorphic approach. On the receiving layer neuron, a sigma decoding computation accumulates the messages and recovers the estimate of the weighted sum as
\begin{align}
    x_\text{est}[t] &= x_\text{est}[t - 1] + s[t]
\end{align}

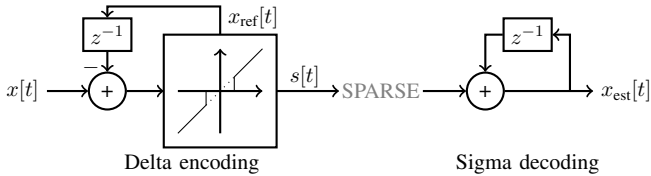
\begin{figure}[!h]
    \centering
    \begin{tikzpicture}[thick, scale=0.75, every node/.style={transform shape}]
    \begin{scope}[shift={(0, 0)}]
        \node (x) at (1, 0) {$x[t]$};
        \node (y) at (6.75, 0) {};
        \node at (6, 0.25) {$s[t]$};
        \node[draw] (z) at (2.5, 1) {$z^{-1}$};
        \node[draw, circle] (sum) at (2.5, 0) {+};
        \draw[->] (x) -- (sum);
        \draw[->] (z) -- (sum) node[pos=0.8, left] {$-$};
        \draw[->] (sum) ++ (2, 1) |- node[pos=0.3, right] {$x_\text{ref}[t]$} ++(-1, 0.5) -| (z);
        \draw[->] (sum) -- ++(1, 0);
        \draw (sum) ++ (1, -1) rectangle ++(2, 2);
        \draw[->] (sum) ++ (3, 0) -- (y);
        \draw[->] (sum) ++ (2, 0) ++(-0.75, 0) -- ++(1.5, 0);
        \draw[->] (sum) ++ (2, 0) ++(0, -0.75) -- ++(0, 1.5);
        \draw[thin] (sum) ++ (2, 0) ++(-0.25, 0) -- ++(0, -0.25) -- ++(-0.5, -0.5);
        \draw[thin] (sum) ++ (2, 0) ++( 0.25, 0) -- ++(0,  0.25) -- ++( 0.5,  0.5);
        \draw[thin, dotted] (sum) ++ (2, 0) ++(-0.25, -0.25) -- ++(0.5,  0.5);
        \node at (4, -1.3) {Delta encoding};
    \end{scope}

    \begin{scope}[shift={(7.95, 0)}]
        \node (x) at (0, 0) {};
        \node[text opacity=0.5] at (-0.6, 0) {SPARSE};
        \node (y) at (3.75, 0) {$x_\text{est}[t]$};
        \node[draw] (z) at (2, 1) {$z^{-1}$};
        \node[draw, circle] (sum) at (1.25, 0) {+};
        \draw[->] (x) -- (sum);
        \draw[->] (sum) -- (y);
        \draw[->] (sum) ++ (1.5, 0) |- (z);
        \draw[->] (z) -| (sum);
        \node at (2, -1.3) {Sigma decoding};
    \end{scope}
\end{tikzpicture}
    \vspace{-6mm}
    \caption{Sparse compression of a temporally redundant signal using delta encoding and its corresponding reconstruction using sigma accumulation.}
    \label{fig:sigma-delta}
\end{figure}

In the ANN conversion method the sigma and delta computation are wrapped around the activation function, typically ReLU. The resulting network is called the Sigma-Delta Neural Network~(SDNN). The sigma-delta logic is illustrated in Figure~\ref{fig:sigma-delta}
.

\noindent\textbf{Loihi 2 inference:} The computational graph of the int SDNN is then laid out on the neurocores of Loihi 2 chip using NxKernel (Intel proprietary software stack). The video frames are saved in additional neurocores for input, which stream the frames as fast as possible showing the true capability of the chip. This mode is called IO unthrottled inference. The region masking module is integrated into the Loihi 2 inference pipeline, as illustrated in Figure \ref{fig:loihi}, to apply masks to the input frames and transmit the masked frames to the YOLO-KP detection network. Additionally, the YOLO-KP computation in each layer is orchestrated in a layer-by-layer fall-through fashion to prioritize the inference latency. While the network is running in IO unthrottled mode, the power and runtime measurements are collected for benchmarking.
\section{Experimental Results}
\subsection{Experimental Setup}
\noindent\textbf{Model:}
The YOLO-KP network is a customized version of Tiny-YOLOv3 network~\cite{adarsh2020yolov3tiny} to be compatible with Loihi~2 architecture (KP refers to Kapoho Point form factor of Loihi 2). It consists of only a single head (the coarser) for prediction. In addition, all the pair of convolution + max pooling layers are collapsed into a single convolution with stride=2 for efficient implementation on Loihi~2. The ANN training, SDNN conversion, and Loihi~2 inference follows the steps described in Sec~\ref{subsec:loihi_inference}. The baseline ANN models are fine-tuned with a static input mask with 50\% sparsity. All region masking results are reported on the models fine-tuned with the static mask. Fine-tuning is performed for 50 epochs using an Adam optimizer with an initial learning rate of $10^{-4}$, cosine learning rate decay, and 4-epoch warmup period. All video frames are rescaled and padded to a uniform size of 448{$\times$}448 before being fed into the YOLO-KP network. 

MGNet consists of a single transformer block with an input size 224{$\times$}224, patch size 16{$\times$}16, and an embedding length of 192. The 448{$\times$}448 input images are downsampled to 224{$\times$}224 and fed into MGNet. The region size is chosen as \si{p}=16 corresponding to the patch size in the transformer block in MGNet.
MGNet is trained for 50 epochs on BDD100K and ImageNetVID, and 100 epochs for KITTI using an Adam optimizer with a learning rate of 0.001. We follow a plateau learning rate decay, that is, the learning rate is reduced by a factor of 0.1 if learning stagnates for more than 5 epochs.

\vspace{3mm}
\noindent\textbf{Dataset:}
We evaluate our approach on three different datasets: KITTI Tracking \cite{Geiger2012CVPR}, BDD100K \cite{Yu_2020_CVPR}, and ImageNet VID \cite{russakovsky2015ImageNet}. KITTI and BDD100K are autonomous driving datasets spanning 9 and 11 categories of objects respectively. KITTI training sequence is split into halves for training and validation, and evaluation is performed on the validation set containing 21 sequences. The results on ImageNet-VID are reported on the validation set on 450 video sequences, with a sequence length of 100. 
The results on BDD100K are reported on 200 test samples, each with a sequence length of 200. 

\vspace{3mm}
\noindent\textbf{Evaluation Metrics:}
Along with the mAP score of the YOLO-KP inference, we compare the inference performance of the 
SDNN models (with various region masking schemes) on Loihi~2 based on the following metrics of interest. Power and runtime are measured from the physical Loihi 2 system.
\begin{itemize}
    \item \textbf{Energy:} the energy used to infer a single frame. For the energy consumption of 
    Loihi~2, we consider the energy consumed by the neurocore compute logic, SRAM and the message IO.
    \item \textbf{Throughput:} the number of frames processed per second.
    \item \textbf{Latency:} the wall-clock time difference between injecting the input frame and obtaining the corresponding system prediction.
    \item \textbf{Energy-delay-product~(EDP):} EDP is a combined metric of the efficiency of processing computed as the energy per frame $\times$ latency of inference. It normalizes the trade-off of running faster and consuming more energy versus running slower and consuming less energy.
    \item \textbf{GOPS/W:} It describes the raw computation operations the system can execute for a unit watt of power. For Loihi~2 SDNNs, we use the dense equivalent GOPS that the ANN computes on a GPU architecture so that the advantage of sparse computation on the neuromorphic hardware is evident.
\end{itemize}
\begin{table*}[!ht]
    \centering
    \scriptsize
    \caption{Performance comparison.}
    \label{table:results}
    \setlength\tabcolsep{3pt} 
    \renewcommand{\arraystretch}{1.5} 
    \let\tmp\ul
    \let\ul\underline
    
    \begin{tabular}{c|l|c|l|r|c|r|r|r|r|r}
        \hline
        \multicolumn{1}{c|}{\multirow{3}{*}{\bf Dataset}} &
        \multicolumn{1}{c|}{\multirow{3}{*}{\bf Network (YOLO-KP)}} &
        \multicolumn{1}{c|}{\multirow{3}{*}{\bf Precision}} &
        \multicolumn{1}{c|}{\multirow{3}{*}{\bf Hardware}} &
        \multicolumn{1}{c|}{\multirow{2}{*}{\parbox{1cm}{\centering\bf Frame Sparsity}}} &
        \multicolumn{1}{c|}{\multirow{3}{*}{\parbox{0.8cm}{\centering{\bf mAP}\\@0.5($\uparrow$)}}} &
        \multicolumn{5}{c}{\textbf{Hardware inference cost per frame}}\\ \cline{7-11}
        &&&&&& \multicolumn{1}{c|}{Energy ($\downarrow$)} & \multicolumn{1}{c|}{Latency ($\downarrow$)} & \multicolumn{1}{c|}{Throughput ($\uparrow$)} & \multicolumn{1}{c|}{EDP ($\downarrow$)}  & \multicolumn{1}{c}{GOPS/W} \\ \cline{7-11}
        &&&& \multicolumn{1}{c|}{($\uparrow$)} && \multicolumn{1}{c|}{Total (mJ)} &\multicolumn{1}{c|}{(ms)} &\multicolumn{1}{c|}{(samples/s)} &\multicolumn{1}{c|}{($\mu$Js)} & \multicolumn{1}{c}{($\uparrow$)} \\ \hline\hline
        \multirow{4}{*}{KITTI}
        & SDNN                            & int8 & Loihi~2$^*$         & $0$    & $\bf{0.2901}$ &    $23.01$ &    $2.29$ &     $436.33$ &     $52.72$ & $89.91$ \\
        & SDNN + Mask$_\text{stat}$       & int8 & Loihi~2$^*$         & $0.58$ &      $0.2579$ &    $\ul{18.72}$ &$\ul{2.02}$& $\ul{495.69}$& $\ul{37.78}$& $11.04$ \\
        & SDNN + Mask$_\text{dyn}$        & int8 & Loihi~2$^*$         & $0.58$ &      $0.2746$ &    $21.75$ &    $2.27$ &     $439.87$ &     $49.45$ & $\ul{95.07}$ \\
        & SDNN + Mask$_\text{stat+dyn}$   & int8 & Loihi~2$^*$         & $0.58$ &      $\ul{0.2792}$ &$\bf{17.07}$& $\bf1.87$ &  $\bf534.33$&  $\bf31.96$ & $\bf{121.11}$ \\
        \hline\hline
         \multirow{4}{*}{BDD100K}
        & SDNN                            & int8 & Loihi~2$^*$         & $0$    & $\bf{0.2335}$ &    $20.14$ &    $2.01$ &     $498.60$ &     $40.40$ & $102.73$ \\
        & SDNN + Mask$_\text{stat}$       & int8 & Loihi~2$^*$         & $0.61$ &      $0.1971$ & $\bf13.44$ & $\bf1.45$ &  $\bf686.60$ &  $\bf19.58$ & $\bf153.90$ \\
        & SDNN + Mask$_\text{dyn}$        & int8 & Loihi~2$^*$         & $0.62$ &      $\ul{0.2269}$ &    $20.79$ &    $2.13$ &     $467.95$ &     $44.43$ & $99.52$ \\
        & SDNN + Mask$_\text{stat+dyn}$   & int8 & Loihi~2$^*$         & $0.61$ &      $0.2255$ &$\ul{16.43}$&$\ul{1.74}$& $\ul{574.76}$& $\ul{28.59}$&$\ul{125.94}$ \\
        \hline\hline
        \multirow{4}{*}{ImageNet-VID}
        & SDNN                            & int8 & Loihi~2$^*$         & $0$    & $\bf{0.2845}$ &    $22.04$ &    $2.18$ &     $459.12$ &     $48.01$ & $94.41$ \\
        & SDNN + Mask$_\text{stat}$       & int8 & Loihi~2$^*$         & $0.42$ &      $0.2743$ &$\bf{19.34}$& $\bf2.06$ &  $\bf484.45$ &  $\bf39.92$ & $\bf{107.59}$ \\
        & SDNN + Mask$_\text{dyn}$        & int8 & Loihi~2$^*$         & $0.43$ &      $0.2771$ &    $21.92$ &    $2.18$ &     $456.99$ &     $47.98$ & $94.90$ \\
        & SDNN + Mask$_\text{stat+dyn}$   & int8 & Loihi~2$^*$         & $0.42$ &      $\ul{0.2791}$ &    $\ul{20.02}$ &$\ul{2.17}$& $\ul{460.01}$& $\ul{45.68}$& $\ul{99.02}$ \\
        \hline
        \multicolumn{11}{p{17.0cm}}{\scriptsize$^*$ The Loihi~2 networks were trained using Lava-dl 0.6.0. Workloads were characterized on an Alia Point 32 chip Loihi~2 system (N3C1 silicon) running on NxCore v2.5.8 and NxKernel v0.1.0 with on-chip IO unthrottled sequencing of input frames.\par
        $^\ddagger$ Performance results are based on testing as of November 2024 and may not reflect all publicly available security updates. Results may vary.
        } \\
    \end{tabular}
    \let\ul\tmp
\end{table*}

\begin{table}[htbp]
    \centering
    \caption{Comparison with ANN}
    \begin{tabular}{c|l|r|c|c}
    \hline
    \bf Dataset & \bf Mask &
        \parbox{1cm}{\centering\bf Frame Sparsity} &
        \parbox{1cm}{\centering{\bf mAP} @0.5($\uparrow$) ANN} &  
        \parbox{1cm}{\centering{\bf mAP} @0.5($\uparrow$) SDNN} \\
        \hline\hline
         \multirow{4}{*}{KITTI} & None & 0 & 0.2938	& 0.2901 \\
         & Mask$_\text{stat}$ & 0.58 & 0.264	& 0.2579 \\
         & Mask$_\text{dyn}$ & 0.58 & 0.2804	& 0.2746\\
         & Mask$_\text{stat+dyn}$ & 0.58 &0.2878	& 0.2792 \\
         \hline
        \multirow{4}{*}{BDD100K} & None & 0 & 0.2437 & 0.2379\\
         & Mask$_\text{stat}$ & 0.61 & 0.2017 & 0.1971\\
         & Mask$_\text{dyn}$ & 0.62 & 0.2326 & 0.2269\\
         & Mask$_\text{stat+dyn}$ & 0.61 & 0.2307 & 0.2255\\
         \hline
          \multirow{4}{*}{ImageNet-VID} & None & 0 & 0.2974 & 0.2845\\       
         & Mask$_\text{stat}$ & 0.42 & 0.2856 & 0.2743\\
         & Mask$_\text{dyn}$ & 0.43 & 0.2908 & 0.2771 \\
         & Mask$_\text{stat+dyn}$ & 0.42 & 0.2924 & 0.2791\\
         \hline
    \end{tabular}
    \label{tab:ann_vs_sdnn}
\end{table}
\vspace{-5mm}

\subsection{Results and Analysis}

Table \ref{table:results} presents mAP@0.5 and hardware inference cost on Loihi~2 for SDNN models, with and without input masks. We further compare masking using static masks ($Mask_{stat}$), dynamic masks ($Mask_{dyn}$), and a combination of both ($Mask_{stat+dyn}$). The hyperparameters $k_{s}$ and $t_{reg}$ are adjusted so as to achieve the same level of input frame sparsity for each masking strategy. We use $k_{s}$=0.2 and $t_{reg}$=0.1 for $Mask_{stat+dyn}$ for all three datasets. We use $t_{reg}$=0.08 for $Mask_{dyn}$ on KITTI and BDD100K and, $t_{reg}$=0.1 for $Mask_{dyn}$ on ImageNetVID. The $k_{s}$ value for $Mask_{stat}$ is set as 1 - frame sparsity (reported in the Table \ref{table:results}).

Static mask performs the best in terms of hardware inference cost providing an improvement in EDP by up to 
2.06$\times$ over baseline SDNN on Loihi~2. This is because a static mask ensures minimum additional computation and data movement between consecutive frames since there are no events arising from the masked regions. However, static mask alone suffers from significant accuracy degradation (3.22\% on KITTI and 3.61\%  on BDD100K over baseline SDNN) and may raise security concerns due to missed objects. On the other hand, although dynamic masking improves detection performance, it cannot achieve significant improvement in hardware inference cost.
Since the SDNN processes delta frames, the dynamic mask incurs additional messages for the disjoint regions of mask in consecutive frames: first for nullifying the messages in previous frame and second for the newly introduced region which results in additional communication overhead.
Combined masking achieves a suitable trade-off between mAP@0.5 and inference cost, achieving an improvement in throughput by 1.22$\times$ and EDP by 1.65$\times$ with a mAP degradation of only 
1.09\% over a baseline SDNN model on KITTI.
For BDD100K, we achieve an improvement in EDP by 1.4{$\times$} with 61\% sparsity 
using our proposed masking strategy with an accuracy degradation of $<$1\%. For ImageNet-VID, we achieve a 1.05$\times$ improvement in EDP with 42\% input frame sparsity,  resulting in a $\sim$0.5\% degradation in mAP. 
Although, inducing an input frame sparsity of $\sim$42\% increases event sparsity by 1.54$\times$ and synaptic sparsity by 1.2$\times$, the energy and latency benefits observed on Loihi are relatively smaller, indicating that a higher input frame sparsity is necessary for achieving EDP improvements on Loihi. 

We demonstrate results for input masking on an ANN model with the same architecture in Table \ref{tab:ann_vs_sdnn}. Combined masking shows promising results on ANN models with a mAP degradation of 0.6\% on KITTI, and 1.3\% on BDD100K for an input sparsity of $\sim$60\% and 0.5\% on ImageNet-VID for an input sparsity of 42\%. SDNN models show performance close to its ANN counterpart, with a maximum degradation 0.5\% for BDD100K, 0.8\% on KITTI, and 1.3\% on ImageNet-VID.

\begin{figure*}
    \centering
    \includegraphics[width=0.7\linewidth]{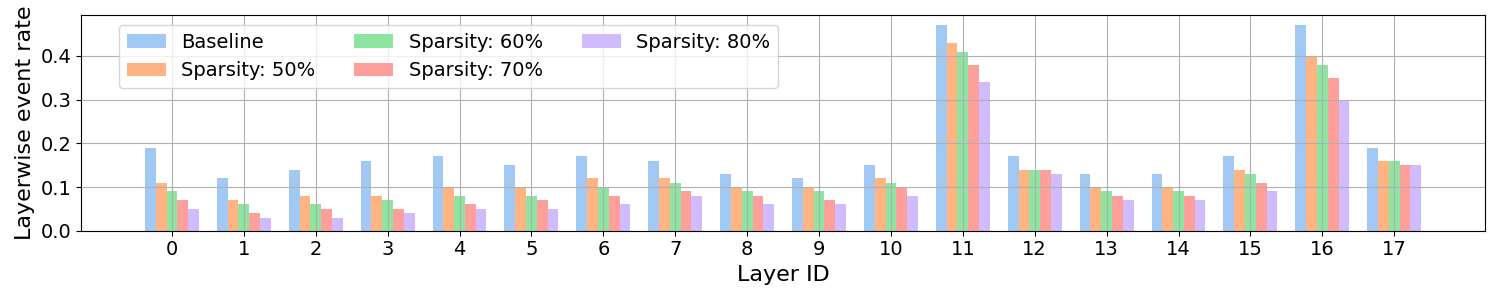}
    \vspace{-3mm}
    \caption{Layerwise event-rate for different input mask sparsity on ImageNet-VID using Tiny-YOLO. This demonstrates how inducing sparsity at the input propagates to the intermediate layers.}
    \label{fig:layerwise}
\end{figure*}

The layerwise event rate for all intermediate layers for SDNN are presented in Figure \ref{fig:layerwise}. We observe a significant drop in event rate across all layers with region masking at the input, with a particularly higher drop in initial layers.

\noindent\textbf{Region Mask Generator:}
The accuracy of prediction of the MGNet with respect to ground truth bounding boxes is evaluated using mIoU scores, given in Table \ref{tab:rmask_precision}. We achieve mIoU scores of 72\%, 67\%, and 62\% on KITTI, BDD100K and ImageNet-VID datasets respectively, with an input resolution of 224$\times$224.
MGNet, with an input resolution of 224$\times$224, has a cost of 0.161 GMACs, which is $\sim$15\% of the YOLO-KP, which involves a computational cost of 1.034 GMACs per sample. 
\begin{table}[htbp]
    \centering
    \caption{mIoU and GFLOPs of MGNet with 224$\times$224 input images}
    \begin{tabular}{c|c|c}
    \toprule
         \textbf{Dataset} & \textbf{mIoU} & \textbf{GFLOPs} \\
         \midrule
          KITTI & 0.72 & 0.161\\
         \midrule
         BDD100K & 0.67 & 0.161\\
        \midrule
         ImageNet-VID & 0.62 & 0.161\\
    \bottomrule
    \end{tabular}
    \label{tab:rmask_precision}
\end{table}
\vspace{-5mm}

\subsection{Qualitative Analysis:}
\begin{figure}
    \centering
    \includegraphics[width=\linewidth]{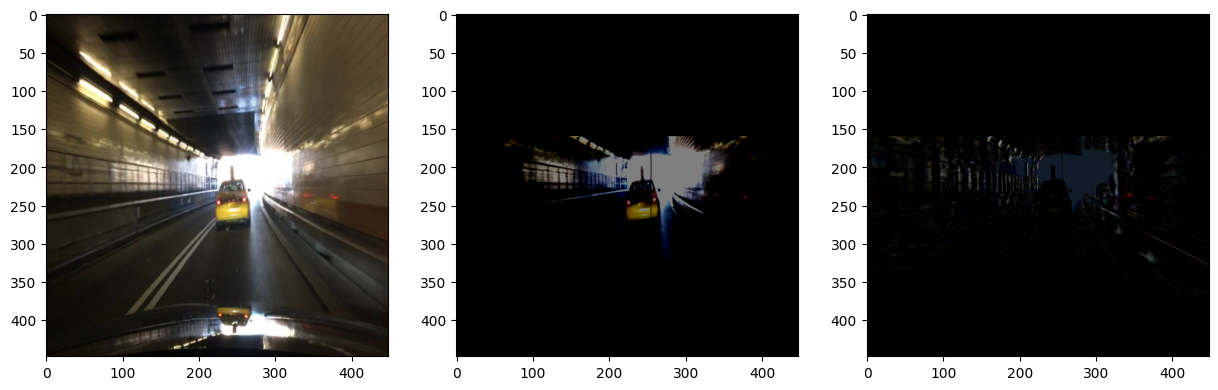}
    \includegraphics[width=\linewidth]{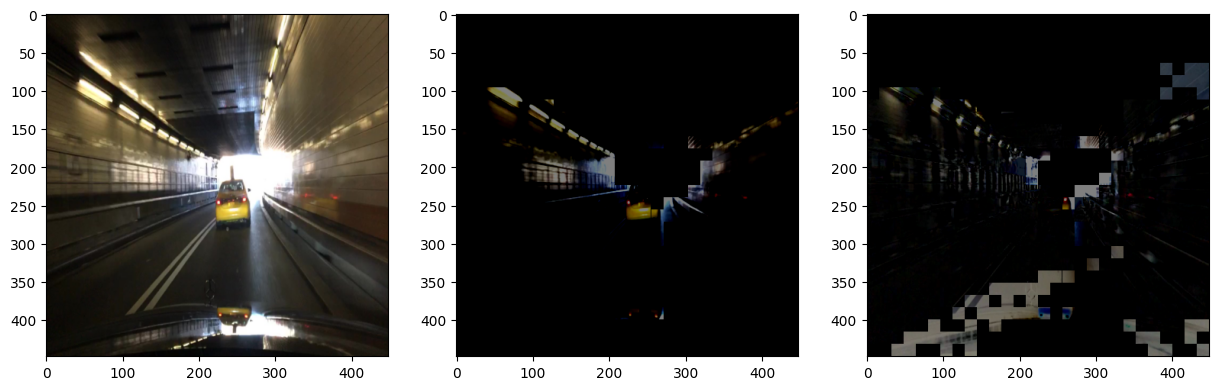}
    \includegraphics[width=\linewidth]{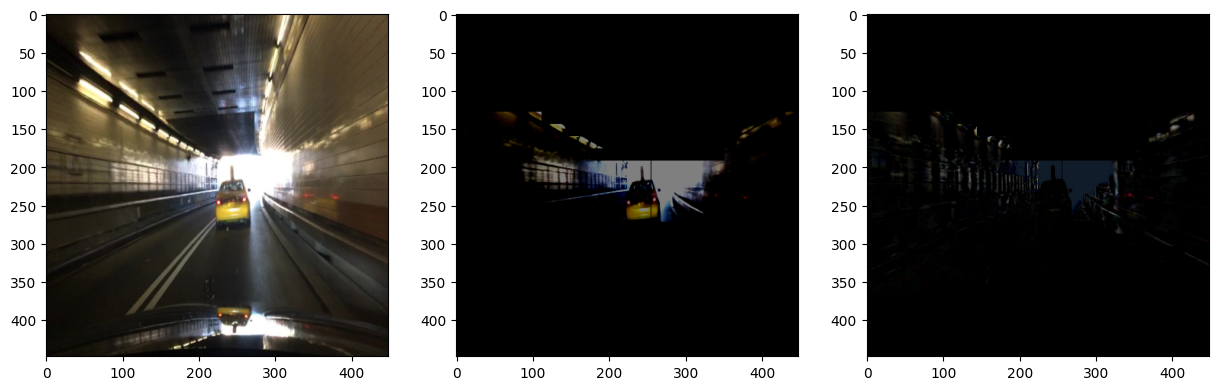}
    \caption{(From left) The input frame before masking, input frame after masking and the delta-encoded frames for static (first row), dynamic (second row) and combined (third row) masking. Dynamic masking shows a few additional patches in the delta-encoded frame (second row right image), which leads to increased computational cost.}
    \vspace{-6mm}
    \label{fig:delta_encoded_frame}
\end{figure}

The masked input frame and the delta-encoded frames for the YOLO-KP SDNN on Loihi are visualized in Figure \ref{fig:delta_encoded_frame}. 
Since the dynamic mask is different for each frame, additional delta patches need to be transmitted, which are observed as the square patches in the delta-encoded frame for dynamic masking. This leads to an increase in computation and traffic in the dynamic case compared to static case, providing limited improvements in the hardware inference cost.

\section{Conclusions}
Sigma-delta networks significantly reduce energy consumption by exploiting temporal redundancy across consecutive frames in a video sequence. However, they still involve redundant computations by processing events regardless of their significance to the task at hand. We propose an input region masking strategy to eliminate events from insignificant regions of an image. A combination of static mask obtained from ground truth object locations in the training set and dynamic mask predicted from the current frame achieves a good trade-off between mAP@0.5 and hardware inference cost. Using our proposed masking strategy, we reduce EDP by 1.65{$\times$} for an SDNN model running on Loihi 2. 
Preliminary analysis of the SDNN inference on Loihi 2 points at inter-chip communication bottleneck which will be improved on future generations promising up to 3.8x improvement in runtime and 9.35x improved EDP. In the future, this masking approach can be extended to eliminate the readout energy for skipped image patches on Loihi 2 by predicting which regions to skip based on prior frames.

\label{sec:conclusion}

\bibliographystyle{IEEEbib}
\footnotesize
\bibliography{ref}

\begin{thebibliography}{10}

\bibitem{he2016deep}
Kaiming He, Xiangyu Zhang, Shaoqing Ren, and Jian Sun,
\newblock ``Deep residual learning for image recognition,''
\newblock in {\em Proceedings of the IEEE conference on computer vision and pattern recognition}, 2016, pp. 770--778.

\bibitem{dosovitskiy2020image}
Alexey Dosovitskiy, Lucas Beyer, Alexander Kolesnikov, Dirk Weissenborn, Xiaohua Zhai, Thomas Unterthiner, Mostafa Dehghani, Matthias Minderer, Georg Heigold, Sylvain Gelly, Jakob Uszkoreit, and Neil Houlsby,
\newblock ``An image is worth 16x16 words: Transformers for image recognition at scale,''
\newblock in {\em International Conference on Learning Representations}, 2021.

\bibitem{russakovsky2015ImageNet}
Olga Russakovsky, Jia Deng, Hao Su, Jonathan Krause, Sanjeev Satheesh, Sean Ma, Zhiheng Huang, Andrej Karpathy, Aditya Khosla, Michael Bernstein, Alexander~C. Berg, and Li~Fei-Fei,
\newblock ``{ImageNet} large scale visual recognition challenge,''
\newblock {\em International Journal of Computer Vision}, vol. 115, no. 3, pp. 211--252, April 2015.

\bibitem{lin2014microsoftcoco}
Tsung-Yi Lin, Michael Maire, Serge Belongie, James Hays, Pietro Perona, Deva Ramanan, Piotr Doll{\'a}r, and C~Lawrence Zitnick,
\newblock ``Microsoft {COCO}: Common objects in context,''
\newblock in {\em Computer Vision--ECCV 2014: 13th European Conference, Zurich, Switzerland, September 6-12, 2014, Proceedings, Part V 13}. Springer, 2014, pp. 740--755.

\bibitem{neuro_frontiers}
Giacomo Indiveri et~al.,
\newblock ``Frontiers in neuromorphic engineering,''
\newblock {\em Frontiers in Neuroscience}, vol. 5, 2011.

\bibitem{dsnn_conversion1}
Yongqiang Cao et~al.,
\newblock ``Spiking deep convolutional neural networks for energy-efficient object recognition,''
\newblock {\em International Journal of Computer Vision}, vol. 113, pp. 54--66, 05 2015.

\bibitem{davies2018loihi}
Mike Davies, Narayan Srinivasa, Tsung-Han Lin, Gautham Chinya, Yongqiang Cao, Sri~Harsha Choday, Georgios Dimou, Prasad Joshi, Nabil Imam, Shweta Jain, et~al.,
\newblock ``Loihi: A neuromorphic manycore processor with on-chip learning,''
\newblock {\em Ieee Micro}, vol. 38, no. 1, pp. 82--99, 2018.

\bibitem{synsense}
``{Xylo-SynSense},'' \url{https://open-neuromorphic.org/neuromorphic-computing/hardware/xylo-synsense/},
\newblock Accessed: 2023-09-30.

\bibitem{furber2020spinnaker}
Steve Furber and Petruț Bogdan,
\newblock {\em Spinnaker-a spiking neural network architecture},
\newblock Now publishers, 2020.

\bibitem{kundu2021spike}
Souvik Kundu, Gourav Datta, Massoud Pedram, and Peter~A Beerel,
\newblock ``Spike-thrift: Towards energy-efficient deep spiking neural networks by limiting spiking activity via attention-guided compression,''
\newblock in {\em Proceedings of the IEEE/CVF winter conference on applications of computer vision}, 2021, pp. 3953--3962.

\bibitem{rathi2018stdp}
Nitin Rathi, Priyadarshini Panda, and Kaushik Roy,
\newblock ``{STDP}-based pruning of connections and weight quantization in spiking neural networks for energy-efficient recognition,''
\newblock {\em IEEE Transactions on Computer-Aided Design of Integrated Circuits and Systems}, vol. 38, no. 4, pp. 668--677, 2018.

\bibitem{shi2019soft}
Yuhan Shi, Leon Nguyen, Sangheon Oh, Xin Liu, and Duygu Kuzum,
\newblock ``A soft-pruning method applied during training of spiking neural networks for in-memory computing applications,''
\newblock {\em Frontiers in neuroscience}, vol. 13, pp. 405, 2019.

\bibitem{lu2020exploring}
Sen Lu and Abhronil Sengupta,
\newblock ``Exploring the connection between binary and spiking neural networks,''
\newblock {\em Frontiers in neuroscience}, vol. 14, pp. 535, 2020.

\bibitem{sorbaro2020optimizing}
Martino Sorbaro, Qian Liu, Massimo Bortone, and Sadique Sheik,
\newblock ``Optimizing the energy consumption of spiking neural networks for neuromorphic applications,''
\newblock {\em Frontiers in neuroscience}, vol. 14, pp. 662, 2020.

\bibitem{shrestha2024efficient}
Sumit~Bam Shrestha, Jonathan Timcheck, Paxon Frady, Leobardo Campos-Macias, and Mike Davies,
\newblock ``Efficient video and audio processing with loihi 2,''
\newblock in {\em ICASSP 2024-2024 IEEE International Conference on Acoustics, Speech and Signal Processing (ICASSP)}. IEEE, 2024, pp. 13481--13485.

\bibitem{Geiger2012CVPR}
Andreas Geiger, Philip Lenz, and Raquel Urtasun,
\newblock ``Are we ready for autonomous driving? the {KITTI} vision benchmark suite,''
\newblock in {\em Conference on Computer Vision and Pattern Recognition (CVPR)}, 2012.

\bibitem{Yu_2020_CVPR}
Fisher Yu, Haofeng Chen, Xin Wang, Wenqi Xian, Yingying Chen, Fangchen Liu, Vashisht Madhavan, and Trevor Darrell,
\newblock ``{BDD100K}: A diverse driving dataset for heterogeneous multitask learning,''
\newblock in {\em Proceedings of the IEEE/CVF Conference on Computer Vision and Pattern Recognition (CVPR)}, June 2020.

\bibitem{orchard2021loihi2}
Garrick Orchard, E~Paxon Frady, Daniel Ben~Dayan Rubin, Sophia Sanborn, Sumit~Bam Shrestha, Friedrich~T Sommer, and Mike Davies,
\newblock ``Efficient neuromorphic signal processing with loihi 2,''
\newblock in {\em 2021 IEEE Workshop on Signal Processing Systems (SiPS)}. IEEE, 2021, pp. 254--259.

\bibitem{panda_res}
Priyadarshini Panda et~al.,
\newblock ``Toward scalable, efficient, and accurate deep spiking neural networks with backward residual connections, stochastic softmax, and hybridization,''
\newblock {\em Frontiers in Neuroscience}, vol. 14, 2020.

\bibitem{spiking_lstm}
Gourav Datta, Haoqin Deng, Robert Aviles, Zeyu Liu, and Peter~A. Beerel,
\newblock ``Bridging the gap between spiking neural networks \& {LSTMs} for latency \& energy efficiency,''
\newblock in {\em 2023 IEEE/ACM International Symposium on Low Power Electronics and Design (ISLPED)}, 2023, vol.~1, pp. 1--6.

\bibitem{lee_dsnn}
Jun~Haeng Lee et~al.,
\newblock ``Training deep spiking neural networks using backpropagation,''
\newblock {\em Frontiers in Neuroscience}, vol. 10, 2016.

\bibitem{Rathi2020Enabling}
Nitin Rathi, Gopalakrishnan Srinivasan, Priyadarshini Panda, and Kaushik Roy,
\newblock ``Enabling deep spiking neural networks with hybrid conversion and spike timing dependent backpropagation,''
\newblock in {\em International Conference on Learning Representations}, 2020.

\bibitem{su2023deep}
Qiaoyi Su, Yuhong Chou, Yifan Hu, Jianing Li, Shijie Mei, Ziyang Zhang, and Guoqi Li,
\newblock ``Deep directly-trained spiking neural networks for object detection,''
\newblock in {\em Proceedings of the IEEE/CVF International Conference on Computer Vision}, 2023, pp. 6555--6565.

\bibitem{kim2021classificationdirectlytrainingspiking}
Youngeun Kim, Joshua Chough, and Priyadarshini Panda,
\newblock ``Beyond classification: Directly training spiking neural networks for semantic segmentation,''
\newblock {\em arXiv preprint arXiv:2110.07742}, 2021.

\bibitem{intelHalaPoint}
``Intel builds world’s largest neuromorphic system to enable more sustainable {AI},'' \url{https://www.intel.com/content/www/us/en/newsroom/news/intel-builds-worlds-largest-neuromorphic-system.html},
\newblock Accessed: 2024-11-11.

\bibitem{liang2022evit}
Youwei Liang, Chongjian Ge, Zhan Tong, Yibing Song, Jue Wang, and Pengtao Xie,
\newblock ``Not all patches are what you need: Expediting vision transformers via token reorganizations,''
\newblock {\em arXiv preprint arXiv:2202.07800}, 2022.

\bibitem{fayyaz2022ats}
Mohsen Fayyaz, Soroush Abbasi~Kouhpayegani, Farnoush Rezaei~Jafari, Eric Sommerlade, Hamid~Reza Vaezi~Joze, Hamed Pirsiavash, and Juergen Gall,
\newblock ``Adaptive token sampling for efficient vision transformers,''
\newblock {\em European Conference on Computer Vision (ECCV)}, 2022.

\bibitem{bolya2022tokenmerge}
Daniel Bolya, Cheng-Yang Fu, Xiaoliang Dai, Peizhao Zhang, Christoph Feichtenhofer, and Judy Hoffman,
\newblock ``Token merging: Your {ViT} but faster,''
\newblock {\em arXiv preprint arXiv:2210.09461}, 2022.

\bibitem{feng2022eye}
Yu~Feng, Nathan Goulding-Hotta, Asif Khan, Hans Reyserhove, and Yuhao Zhu,
\newblock ``Real-time gaze tracking with event-driven eye segmentation,''
\newblock in {\em 2022 IEEE Conference on Virtual Reality and 3D User Interfaces (VR)}. IEEE, 2022, pp. 399--408.

\bibitem{reidy2024hirise}
Brendan Reidy, Sepehr Tabrizchi, Mohamadreza Mohammadi, Shaahin Angizi, Arman Roohi, and Ramtin Zand,
\newblock ``{HiRISE}: High-resolution image scaling for edge ml via in-sensor compression and selective roi,''
\newblock {\em arXiv preprint arXiv:2408.03956}, 2024.

\bibitem{sarkar2024maskvd}
Sreetama Sarkar, Gourav Datta, Souvik Kundu, Kai Zheng, Chirayata Bhattacharyya, and Peter~A Beerel,
\newblock ``{MaskVD}: Region masking for efficient video object detection,''
\newblock {\em arXiv preprint arXiv:2407.12067}, 2024.

\bibitem{Wang_2023_ICCV_masked_spikformer}
Ziqing Wang, Yuetong Fang, Jiahang Cao, Qiang Zhang, Zhongrui Wang, and Renjing Xu,
\newblock ``Masked spiking transformer,''
\newblock in {\em Proceedings of the IEEE/CVF International Conference on Computer Vision (ICCV)}, October 2023, pp. 1761--1771.

\bibitem{kaiser2024energy}
Md~Abdullah-Al Kaiser, Sreetama Sarkar, Peter~A Beerel, Akhilesh~R Jaiswal, and Gourav Datta,
\newblock ``Energy-efficient \& real-time computer vision with intelligent skipping via reconfigurable {CMOS} image sensors,''
\newblock {\em arXiv preprint arXiv:2409.17341}, 2024.

\bibitem{adarsh2020yolov3tiny}
Pranav Adarsh, Pratibha Rathi, and Manoj Kumar,
\newblock ``Yolo v3-tiny: Object detection and recognition using one stage improved model,''
\newblock in {\em 2020 6th International Conference on Advanced Computing and Communication Systems (ICACCS)}, 2020, pp. 687--694.

\end{thebibliography}


\begin{thebibliography}{00}
\bibitem{b1} G. Eason, B. Noble, and I. N. Sneddon, ``On certain integrals of Lipschitz-Hankel type involving products of Bessel functions,'' Phil. Trans. Roy. Soc. London, vol. A247, pp. 529--551, April 1955.
\bibitem{b2} J. Clerk Maxwell, A Treatise on Electricity and Magnetism, 3rd ed., vol. 2. Oxford: Clarendon, 1892, pp.68--73.
\bibitem{b3} I. S. Jacobs and C. P. Bean, ``Fine particles, thin films and exchange anisotropy,'' in Magnetism, vol. III, G. T. Rado and H. Suhl, Eds. New York: Academic, 1963, pp. 271--350.
\bibitem{b4} K. Elissa, ``Title of paper if known,'' unpublished.
\bibitem{b5} R. Nicole, ``Title of paper with only first word capitalized,'' J. Name Stand. Abbrev., in press.
\bibitem{b6} Y. Yorozu, M. Hirano, K. Oka, and Y. Tagawa, ``Electron spectroscopy studies on magneto-optical media and plastic substrate interface,'' IEEE Transl. J. Magn. Japan, vol. 2, pp. 740--741, August 1987 [Digests 9th Annual Conf. Magnetics Japan, p. 301, 1982].
\bibitem{b7} M. Young, The Technical Writer's Handbook. Mill Valley, CA: University Science, 1989.
\bibitem{b8} D. P. Kingma and M. Welling, ``Auto-encoding variational Bayes,'' 2013, arXiv:1312.6114. [Online]. Available: https://arxiv.org/abs/1312.6114
\bibitem{b9} S. Liu, ``Wi-Fi Energy Detection Testbed (12MTC),'' 2023, gitHub repository. [Online]. Available: https://github.com/liustone99/Wi-Fi-Energy-Detection-Testbed-12MTC
\bibitem{b10} ``Treatment episode data set: discharges (TEDS-D): concatenated, 2006 to 2009.'' U.S. Department of Health and Human Services, Substance Abuse and Mental Health Services Administration, Office of Applied Studies, August, 2013, DOI:10.3886/ICPSR30122.v2
\bibitem{b11} K. Eves and J. Valasek, ``Adaptive control for singularly perturbed systems examples,'' Code Ocean, Aug. 2023. [Online]. Available: https://codeocean.com/capsule/4989235/tree
\end{thebibliography}

\end{document}